\documentclass[lettersize,journal]{IEEEtran}
\usepackage{amsmath,amsfonts}
\usepackage{algorithmic}
\pdfoutput=1
\usepackage{array}
\usepackage[caption=false,font=normalsize,labelfont=sf,textfont=sf]{subfig}
\usepackage{textcomp}
\usepackage{stfloats}
\usepackage{url}
\usepackage{verbatim}
\usepackage{graphicx}
\usepackage{utfsym}
\usepackage{titlesec}
\usepackage{booktabs}
\usepackage{CJKutf8}
\usepackage{setspace}
\def\BibTeX{{\rm B\kern-.05em{\sc i\kern-.025em b}\kern-.08em
    T\kern-.1667em\lower.7ex\hbox{E}\kern-.125emX}}
\usepackage{balance}
\begin{document}
\title{Towards Training A Chinese Large Language Model for Anesthesiology}
\begin{comment}
\author{
Zhonghai Wang,~\IEEEmembership{Member,~IEEE, }
Yibing Zhan,~\IEEEmembership{Member,~IEEE, }
Bohao Zhou,~\IEEEmembership{Member,~IEEE, }
Yanhong Li,~\IEEEmembership{Member,~IEEE, }
Chong Zhang,~\IEEEmembership{Member,~IEEE, }
Liang Ding,~\IEEEmembership{Member,~IEEE, }
Hua Jin,~\IEEEmembership{Member,~IEEE, }
Jun Peng,~\IEEEmembership{Member,~IEEE, }
Weifeng Liu,~\IEEEmembership{Member,~IEEE, }
Dapeng Tao,~\IEEEmembership{Member,~IEEE, }
\end{comment}
\author{
Zhonghai Wang*,
Jie Jiang*,
Yibing Zhan,
Bohao Zhou,
Yanhong Li,
Chong Zhang,
Liang Ding,
Hua Jin,\\
Jun Peng,
Xu Lin,
and Weifeng Liu
\thanks{Zhonghai Wang, Bohao Zhou and Weifeng Liu are with the College of Control Science and Engineering,China University of Petroleum (East China),Qingdao 266580, China.

Jie Jiang is with Tencent.

Yibing Zhan is with JD Explore Academy.

Yanhong Li and Chong Zhang are with Yunnan University.

Liang Ding is with Zhejiang University.

Hua Jin and Jun Peng are with First People's Hospital of Yunnan Province.

Xu Lin is with Yunnan United Vision Technology Co., Ltd

* Zhonghai Wang and Jie Jiang contributed equally to this manuscript.

%\# Corresponding Author: Yibing Zhan. Email: zybjy@mail.ustc.edu.cn
}}
\markboth{Journal of \LaTeX\ Class Files,~Vol.~14, No.~8, August~2021}%
{Shell \MakeLowercase{\textit{et al.}}: A Sample Article Using IEEEtran.cls for IEEE Journals}

%\IEEEpubid{0000--0000/00\$00.00~\copyright~2021 IEEE}

\maketitle

\begin{abstract}
Medical large language models (LLMs) have gained popularity recently due to their significant practical utility. However, most existing research focuses on general medicine, and there is a need for in-depth study of LLMs in specific fields like anesthesiology.  To fill the gap, we introduce Hypnos, a Chinese Anesthesia model built upon existing LLMs, e.g., Llama.   Hypnos' contributions have three aspects:  
1) The data, such as utilizing Self-Instruct, acquired from current LLMs likely includes inaccuracies.  Hypnos implements a cross-filtering strategy to improve the data quality. This strategy involves using one LLM to assess the quality of the generated data from another LLM and filtering out the data with low quality. 2) Hypnos employs a general-to-specific training strategy that starts by fine-tuning LLMs using the general medicine data and subsequently improving the fine-tuned LLMs using data specifically from Anesthesiology. The general medical data supplement the medical expertise in Anesthesiology and enhance the effectiveness of Hypnos' generation.  3) We introduce a standardized benchmark for evaluating medical LLM in Anesthesiology.  Our benchmark includes both publicly available instances from the Internet and privately obtained cases from the Hospital.  Hypnos outperforms other medical LLMs in anesthesiology in metrics, GPT-4, and human evaluation on the benchmark dataset.
\end{abstract}


\section{Introduction}
\IEEEPARstart{I}{n} recent years, large language models (LLMs), such as chatGPT \cite{open2023introducing}, have attracted significant interest due to their exceptional performance in understanding instructions and generating human-like responses in a wide range of domains. 
Furthermore, thanks to the open-source foundational language models, including Llama \cite{touvron2023llama}, Llama 2 \cite{touvron2023llama2}, and Bloom \cite{scao2022bloom}, leveraging open-source LLMs and adapting them to specific applications/domains, especially medicine, have been popular. 

Considering the potential practical values, numerous medical LLMs have been proposed in the literature. For example, Med-PaLM \cite{singhal2023large} and Med-PaLM2 \cite{singhal2023towards} are AI tools designed to answer questions about medical information based on PaLM \cite{chowdhery2022palm}. BenTsao \cite{wang2023huatuo} is a Llama-based model that has been supervised-fine-tuned with generated Question-Answer (Q\&A) instances. HuatuoGPT \cite{huatuogpt-2023} is a biomedical chatbot incorporating distilled data generated by GPT-3.5-turbo and obtained from the real world.
However, most of the above medical LLMs are built based on general medical knowledge, and the study of learning an LLM in a specific medical field is relatively rare. 
To our knowledge, there is only one large language model for radiology \cite{liu2023radiology} based on instruction tuning from an existing large-scale dataset: MIMIC-CXR \cite{johnson2019mimic}. Nevertheless, not all specific medical fields have sufficient expertise data, and more studies of learning an LLM in a specific medical field are still needed urgently. 

To mitigate the gap, we present Hypnos, a fine-tuned chat model for Anesthesiology, a fundamental, significant, yet challenging branch of medicine. Doctors can retrieve fundamental anesthesia knowledge through our Hypnos, and patients can talk with Hypnos to obtain solutions to their diseases. The contributions of Hypnos are summarised as follows:

\begin{itemize}
\item 
First, obtaining sufficient and high-quality data in Anesthesiology is difficult. We collected 8,000K pairs of Chinese Q\&A from the freely available data on the Internet, whereas only 18K pairs contain related information on Anesthesiology. A commonly-used solution is to obtain data using self-instruct \cite{wang2022self}
from existing LLMs. However, we found that some of the generated data contained noises. 
Therefore, we design a cross-filtering strategy to obtain high-quality data. Specifically, Hypnos collects data from multiple LLMs, requires one LLM to score the quality of the generating data from another LLM, and drops the generating data with low-quality scores. 

\item Second, knowledge of general medicine may benefit the understanding of Anesthesiology, and more Q\&A data can guarantee the generation capabilities of a medical LLM. However, directly combining data from general medicine and Anesthesiology to tune an LLM would lead to the knowledge from Anesthesiology being overwhelmed by the large-scale general medicine data. Therefore, Hypnos adopts a general-to-specific training strategy that first tunes LLMs using data from the general medical domain and then refines the tuned LLM using the data from Anesthesiology. The benefit of general medicine data is properly introduced to Hypons.

\item Third, datasets for testing medical LLM capabilities are valuable, but there is no standard dataset for testing the medical performance of LLMs in Anesthesiology. Therefore, we develop a standard benchmark for evaluating the performance of LLMs for Anesthesiology. Our benchmark contains 5,900 multiple-choice questions, 556 Q\&A about basic knowledge of anesthesia from the Internet and Books, and 50 real anesthesia cases legally obtained from the private data of the Hospital. % evaluation Three types
\end{itemize}

Experimental results, compared with state-of-the-art medical LLMs, including BenTsao \cite{wang2023huatuo}, ChatMed \cite{zhu2023ChatMed}, QiZhenGPT \cite{RN09}, BianQue-2 \cite{chen2023bianque1}, HuatuoGPT \cite{huatuogpt-2023}, and a general-purpose model, Baichuan-7B \cite{RN10}, demonstrate the superiority of our Hypnos that it obtained the best performance on the dataset of Anesthesiology, no matter metrics evaluation, GPT evaluation and, human evaluation from three aspects: usefulness, harmfulness, and redundancy. 

%加载图片
\begin{figure*}
  \centering
  \includegraphics[width=1\textwidth]{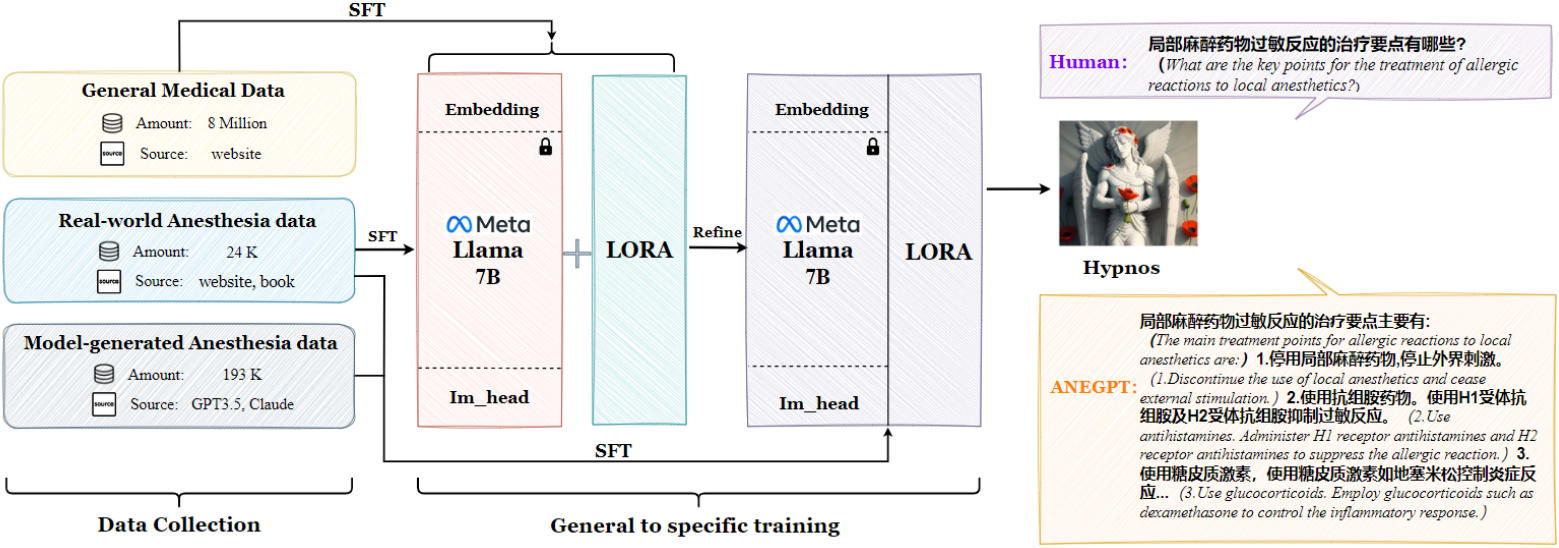}
  \caption{The brief training process of our Hypnos. The training process contains two aspects: data collection and model training. In the data collection, we provide the data types, sizes, and sources. In the model training processes, Hypnos first extends the Chinese Vocabulary, then tunes the Llama with LORA using the general medical data, and last, fully refines the model using anesthesia instruction data. The Lock refers to the component that is fixed and not learnable.}
  \label{fig-1}
\end{figure*}

\section{Related Work}
\subsection{Large Language Models} 
With the development of natural language processing (NLP), large language models (LLMs) have emerged and have revolutionized the field of NLP. A series of LLMs with excellent performance has attracted increasing attention by their powerful language understanding ability, such as GPT-3.5-turbo, GPT-4 \cite{gpt42023}, Claude \cite{Claude}, PaLM \cite{chowdhery2022palm}, and PaLM-2 \cite{anil2023palm}. In addition to the powerful commercial models, various alternative open-source models have gradually emerged, such as Llama \cite{touvron2023llama}, Llama 2 \cite{touvron2023llama2}, Baichuan-7B \cite{RN10}, ChatGLM \cite{du2021glm}, and Bloom \cite{scao2022bloom}. These open-source models are accessible and reproducible, which significantly promotes the development of applying LLMs to different fields. 

Llama is one of the most representative open-source models, with parameter sizes ranging from 7 billion to 65 billion. Based on Llama, a series of fine-tuning models appeared, including Alpaca \cite{taori2023alpaca} and Vicuna \cite{chiang2023vicuna}. Original Llama's performance on several English tasks is comparable to GPT-3.5-turbo, but its performance on Chinese tasks is relatively poor. Therefore, tuning Llama using Chinese is also a hot topic. For instance, Chinese-LLaMA-Alpaca \cite{chinese-llama-alpaca} %github
expands the vocabulary based on Llama with Chinese. The work also inspired other general medical models, such as ChatMed. 
We also borrowed some ideas from Chinese-LLaMA-Alpaca to train our Chinese medical LLM.

\subsection{Medical Large Language Models} 
Applying Large Language Models to medicine is popular. For instance, Chatdoctor \cite{li2023chatdoctor} adapts LLM to the biomedical domain by fine-tuning Llama \cite{touvron2023llama} using conversational demonstrations synthesized by ChatGPT. Baize \cite{xu2023baize} generates multiple rounds of dialogue data through ChatGPT and uses ChatGPT to select the optimal response of the model for feedback learning. Earlier Medical LLMs were English, and recently, there have been a lot of Chinese Medical LLMs. For example, ChatMed \cite{zhu2023ChatMed} uses real inquiry data with ChatGPT to generate responses since the real answers contain a lot of noise. BenTsao \cite{wang2023huatuo} generates dialogue inquiry data based on the medical knowledge base. Besides, BenTsao \cite{wang2023huatuo} uses professional papers as prompts for ChatGPT to generate more professional multiple rounds of dialogue data. HuatuoGPT \cite{huatuogpt-2023} uses not only real data and distillation data from other models but also artificial feedback in training to enhance the performance of the medical field. Nonetheless, the above-mentioned medical LLMs are proposed based on general medical knowledge. To our knowledge, learning an LLM for a specific medical field is still rare, and only one work has explored applying LLMs to a specific medical field: Radiology-GPT \cite{liu2023radiology}%
, an LLM for Radiology based on an existing large-scale radiology dataset. This paper focuses on learning an LLM for Anesthesiology, which contains limited publicly available training data.

\section{Hypnos}
This section presents the details of Hypnos. As shown in Figure \ref{fig-1}, the overview of Hypnos includes Data Collection and Training Details, which will be sequentially explained in the following sections. Besides, we will introduce the Anesthesia LLM Evaluation of Hypnos.

\subsection{Data Collection}
There are no publicly accessible large-scale datasets in Anesthesiology.  We collect the training data from two aspects: 1) real-world data obtained from the Internet and Books and 2) model-generated data obtained from existing LLMs. 

\subsubsection{Real-world Data Obtained from Internet and Books}
Following HuatuoGPT, we first obtain Chinese Q\&A pairs that are freely available on medical websites. The raw data from the Internet is noisy and contains private information. Therefore, we first ignore the data with duplicate information, Garbled characters, and private information, including humans'/Hospitals' names, phone numbers, and addresses. Then, we drop sentences that contain less than three words, which are less informative. In such a manner, the data of general medicine are reduced from 13,000k to 8,000k. Then, we use a keyword-matching strategy to select data that contain keywords related to Anesthesiology. After the keyword-matching, only 18k data are maintained.
%anesthesia data can be extracted from 800k general medical data.
Besides, we also collect 6k Q\&A  data from books as a complementary of the data from the Internet. Specifically, we directly extract short-answer questions and case analyses from books that contain Q\&A. %有问题
Finally, 24K data Obtained from the Internet and Books are obtained as the final real-world data in Anesthesiology.

Here, the keywords include anesthesia, general anesthesia, local anesthesia, anesthetics, propofol, etomidate, and midazolam, and the books include Miller's Anesthesiology, Clinical Anesthesiology, Anesthesia Management, and Anesthesia Papers Collection. More keywords and books can be found in Appendix \ref{Appendix A} and Appendix \ref{Appendix B} of the supplementary materials.

\begin{figure*}[t]
\centering
\includegraphics[width=1\textwidth]{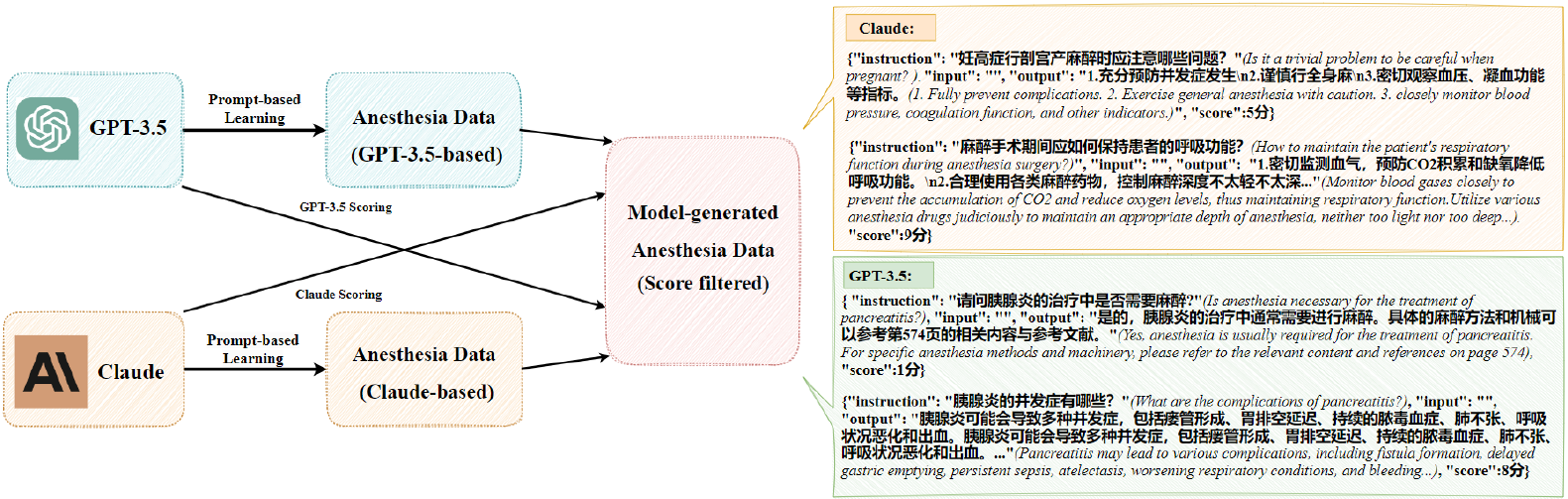}
\caption{Step 1: Use GPT-3.5-turbo and Claude to generate anesthesia-related instruction data. Step 2: Clean up data with obvious incorrectness. Step 3: Use Claude to score the data generated by GPT-3.5-turbo, and use GPT-3.5-turbo to score the data generated by Claude. Step 4: Clean up instruction data with low scores to obtain the final model generation data.}
\label{fig-2}
\end{figure*}

\subsubsection{Model-generated Data Obtained From Existing LLMs}
%量比较小，self-instruct，%
The anesthesia data obtained from the Internet and Books is small-scale. %Following,
Therefore, we adopt the commonly used self-instruct strategy %缺参考文献
to obtain more training data from existing LLMs. Nevertheless, we found that the generated data contain noises, and therefore, as shown in Figure \ref{fig-2}, we propose a cross-filtering strategy to obtain high-quality training data. Some examples with high scores and low scores can be found in Figure \ref{fig-2}. %The details are as follows.
%更具体

First, we designed a prompt for LLMs to generate data and manually selected 30 anesthesia questions and answers from the New Youth Anesthesia website as the seed pool. This prompt can be found in Appendix \ref{Appendix C}.
Each time, one pair of questions and answers is randomly selected from the seed pool as an example input to the self-Instruct strategy. Besides, to make the data generated by the self-Instruct more professional, an anesthesia text is also sent to the LLMs. The anesthesia text is randomly segmented from anesthesia medical books with a length of 400-500 words. 
The LLM will be required to ask questions based on the given anesthesia text following the input anesthesia Q\&A examples. Answers are subsequently generated by the same LLM based on the generated questions and the given anesthesia text.

In order to complete cross-filtering, we adopt GPT-3.5-turbo and Claude to generate data. We choose these two capable LLMs, because Claude is free and GPT-3.5-turbo is relatively cheap. To obtain diverse data, we calculate the Rouge-1 value for each instruction data by comparing it with the preceding one hundred instruction data. The data with a Rouge-1 value greater than 0.5 will be deleted. Besides, we remove all data with the length of question or answer less than ten characters. We spent a week obtaining 110k and 160k raw anesthesia data by GPT-3.5-turbo and Claude, respectively.

The generated data still contains noises. Therefore, we designed another type of prompt to enable LLMs to have the ability to evaluate the quality or the correctness of the generated data. This prompt can be found in Appendix \ref{Appendix D}. %
Since the data generated from one LLM is generally considered high-quality by itself, we require one LLM to score the quality of data generated by another LLM. The quality score is evaluated by considering five aspects: coherence, professionalism, usefulness, harmfulness, and friendliness.
% Let the model score the generated data comprehensively from these 5 aspects.  Clean the data based on different scores, utilize this data to train using the llama model as a foundation, evaluate various models on the test set, and choose the best-cleaned data.
The model takes into account five aspects and assigns a score ranging from 0 to 10, in which a higher score indicates better data quality. We filter out data with lower scores and maintain 193k model-generated data.
The scores obtained from an LLM may still contain bias; one of our future works is to improve the filtering strategy by obtaining a more comprehensive judgment from LLMs. Examples of ratings can be found in the appendix \ref{Appendix E}.
% During the cleaning process, there is no guarantee of 100\% accurate data differentiation; only a certain degree of differentiation can be achieved. 

% More details of data collection, including the prompts, can be found in the supplementary materials \ref{Appendix C}.

\subsection{Training Details}
Hypnos is obtained based on Llama. 
Specifically, Hypnos first extends the vocabulary of Llama, finetunes Llama using LORA with 8,000K general medical data, and fully refines the model using 
217k anesthesia data. %数据量是多少

\subsubsection{Chinese Vocabulary Extension} 
To enhance the efficiency of Chinese encoding, we extend the original vocabulary of Llama by adding more Chinese vocabulary \cite{cui2023efficient}. Specifically, we first use SentencePiece 
 \cite{Kudo2018SentencePieceAS} with Byte-Pair Encoding (BPE) word segmentation to obtain new Chinese vocabulary based on our collected data. Then, we initialize all new tokens with UTF-8 \cite{komatsuzaki2022sparse}. Here, each Chinese word can be represented by three UTF-8 characters. The initialized embedding is obtained by fusing the vectors of the UTF-8 characters with a weight set as the order of the character to identify the Chinese word. Therefore, each Chinese token has a unique initialization.

\subsubsection{General-to-Specific Training Strategy}
%Genral Lora
%Spefict 
% In order to utilize the knowledge of all collected data, we design a general-to-specific training strategy. 
There are 8,000k general medical data and 217K anesthesia data. We validate that knowledge from general medicine may benefit the understanding of Anesthesiology, and more Q\&A data can guarantee the generation capabilities of a medical LLM. We propose a general-to-specific training strategy.

Specifically, for the general training, we use the 8,000k general medical data to train Llama with LORA with one epoch.
We first train the word embedding layer and Imhead output layer while fixing other networks using 1,000K general medical data. 
Then, we adopt LORA adaptation \cite{hu2021lora} in all linear layers except for the word embedding layer and Imhead output layer for time effeciency. We train the parameters of LORA using the remaining 7,000k general medical data. It took eight days on a server with 8 Nvidia A100 40GB GPUs to finish the training.

For the specific training, We further improved our model using the 217k anesthesia medical dataset. We train the model with three additional epochs to obtain the final model. In specific training, all parameters in the model are learnable to achieve better performance.

% to obtain our large anesthesia medical model.
\subsection{Anesthesia LLM Evaluation}
This subsection presents the details of our Anesthesia LLM Evaluation. 

\subsubsection{Anesthesia Evaluation Dataset} Our evaluation dataset contains three parts. %First, 
% In addition to the lack of a large language model specific to anesthesiology, there is also a lack of anesthesia test sets for evaluation. Real data is used to build a test set ANE. The ANE test set is divided into three parts.
The first part is a test set of Anesthesia Q\&A (AneQA), consisting of 556 basic knowledge  Q\&A obtained from online health consultancy websites and professional anesthesia medical books. 
The second part is a test set of Anesthesia Choice Questions (AneCQ), consisting of more than 5,900 anesthesia single-choice questions, which are obtained from the anesthesia professional examinations. %大部分？那么还有一部分来自于？
The third part is a test set of Anesthesia Real Cases (AneRC) that are composed of 50 real anesthesia cases obtained from real patient cases. The real cases do not have professional ground truth.% due to privacy. %需不需要把医院写出来
%50 cases of AneRC contain manually annotated answers from anesthesiologists.

\subsubsection{Evaluation of Metrics} For AneQA, commonly used evaluation metrics, including BLEU \cite{papineni2002bleu}, ROUGE \cite{chin2004rouge}, GLEU \cite{mutton2007gleu}, and Distinct \cite{li2015diversity}, are used to measure the performance of testing methods following \cite{huatuogpt-2023}. Here, BLEU and ROUGE are used to calculate the similarity between the model output and the ground-truth answer, GLUE is used to calculate the fluency of the model output, and Distinct is used to calculate the diversity of the model outputs.  %需要citations %Although automatic evaluation cannot accurately measure the performance of a model, it can roughly represent the quality of the model.

\subsubsection{Manual Evaluation} We also ask anesthesiologists to judge the performance of testing methods. Specifically, we randomly selected 50 data samples from the AneQA test set and all data samples of the AneRC test set for evaluation. Three anesthesiologists participated in the evaluation. Each answer of LLMs is judged from the following aspects:

\begin{itemize}
\item Usefulness assessment: Evaluate whether the model's answers can solve the problems. Consider whether the answer contains necessary information and whether substantive suggestions regarding anesthesia plans or procedures can be provided.

\item Harmfulness assessment: Check whether there are serious logical errors or inaccurate information in the model response. Pay special attention to content that may lead to incorrect medical decisions. Determine if any model answers involve false medical information or hallucinatory suggestions. These may pose serious risks to the patient's health.

\item Redundancy assessment: The proportion of practical and useful parts in the responses generated by the model in the entire response. This can reflect whether the model can focus on the necessary information to solve the problem when generating content, thereby improving efficiency and reducing redundancy.
\end{itemize}
The anesthetist ranks the replies of all comparing models based on the above three aspects.

\subsubsection{GPT-4-based Evaluation} %We continued to randomly extract 50 data samples from the basic knowledge question-answering test set (ANEQA) and another 50 data samples from the case analysis test set (ANERC) for evaluation. %这个和上面是一个东西么？
We also rely on GPT-4 to evaluate the 100 questions based on the three criteria, just as in the manual evaluation. 
% We choose the format of a competition to compare our model with other medical models. 
Specifically, we conduct pair comparison experiments to ask the GPT-4 to select a better one from our method's results and another competitor's. We note that the
evaluation of GPT-4 may be affected by the location of the answers \cite{wang2023large}. To avoid the positional effect, we present answers from different LLMs in a random order to GPT-4. %The identifier in the prompt may have an impact on the result, so it is also randomized.
In order to eliminate bias introduced by the model answer identifiers, such as 'model\_A's response' vs. 'model\_B's response,' we randomly selected half of the data and swapped their identifiers. When given to GPT-4, all methods' names are anonymous. The evaluation prompts and details can be found in the appendix \ref{Appendix E}.

\section{Experiments}

\subsection{Implementation Details}
We used Belle's program \cite{BELLE}%
to fine-tune Llama\_7B with a server equipped with 8 Nvidia A100 40GB GPUs. We used the DeepSpeed library's Zero Stage 3 and opted for offload to load the model parameters and optimizer onto the CPU. The learning rate, batch size, and maximum context were set to 5e-5, 192, and 1024, respectively. All models were trained at most in three epochs. The performance of each epoch was recorded, and the best was used as the final output.

\subsection{Baselines}
%我们的hypnos模型分别与ChatMed、BenTsao、Qizhen、HuaTuo_7B、BianQue、Baichuan-7B_7B模型进行比较。在ANEQA和ANEEC上进行自动评估测试。在ANEQA和ANERC上进行GPT4评估和人工评估。
Six LLMs are used for the performance comparison: ChatMed \cite{zhu2023ChatMed}, BenTsao \cite{wang2023huatuo}, QiZhenGPT \cite{RN09}, HuatuoGPT \cite{huatuogpt-2023}, BianQue-2 \cite{chen2023bianque1}, and Baichuan-7B \cite{RN10}. %
These six LLMs are selected because they perform well in medical Q\&A, and they release their codes on the Internet. In order to ensure the accuracy of the testing, we conducted three inferences on the AneQA and took the average.

\begin{table*}
  \centering
  \caption{Automated evaluation results of medical LLMS on the AneQA.}
  \resizebox{1\textwidth}{!}
  {\begin{tabular}{cccccccccccc}
    \toprule
    years & Model & BLEU-1 & BLEU-2 & BLEU-3 & BLEU-4 & GLEU & ROUGE-1 & ROUGE-2 & ROUGE-L & Distinct-1 & Distinct-2 \\
    \midrule
               & HuatuoGPT & \textbf{21.34} & 6.72  & 2.57 & 1.20 & 8.37  & 18.23& 3.30 & 14.45 & 50.13& 76.47   \\
               & BianQue-2 & 21.01 & 6.56 & 2.47 & 1.17 & 8.41 & 20.32 & 3.42 & 14.90 & 45.47  & 68.67 \\
             & QiZhenGPT & 18.84 & 6.60 & 2.86 & 1.46 & 8.14 & 20.46 & 4.25  &13.77 & 39.27 & 57.70  \\
     2023 & Baichuan-7B & 16.21 & 4.43 & 1.39 & 0.57 & 6.35 & 16.65 & 1.91  & 9.55 & 65.96 & 92.26  \\
               & ChatMed & 9.97 & 2.34 & 0.72 & 0.30 & 4.83 & 14.69 & 1.33  & 10.04 & \textbf{80.39} & \textbf{95.52}  \\
               & Bentsao & 3.50 & 1.52 & 0.78 & 0.48 & 4.05 & 13.83 & 2.78  & 9.55 & 71.37 & 77.49  \\
               &Hypnos(our) &19.34&\textbf{7.17} & \textbf{3.1}  & \textbf{1.64}& \textbf{9.15} & \textbf{21.73} & \textbf{4.69} & \textbf{14.79}  &54.31 & 76.94   \\
    \bottomrule
  \end{tabular}}
  \label{tab:1}
\end{table*}

\begin{table*}
  \caption{Automated evaluation results of medical LLMs on the anesthesia test set AneCQ. "Random" represents randomly selected.}
  \centering
  \resizebox{1\textwidth}{!}
  {\begin{tabular}{ccccccccc}
    \toprule
    Models & Random &HuatuoGPT & BianQue-2.0 & QiZhenGPT & Baichuan-7B & ChatMed & Bentsao & Hypnos\\
    \midrule
    % random & 20.00 & 20.00 & 20.00 & 20.00 & 20.00 & 20.00 & 20.00  \\
    Score& 20.00  & 24.54 & 20.54 & 21.6 & 24.1 & 19.60  & 19.74 & \textbf{25.17} \\   
    \bottomrule
  \end{tabular}}
  \label{tab:2}
\end{table*}

\begin{figure*}[t]
\centering
\includegraphics[width=1\textwidth]{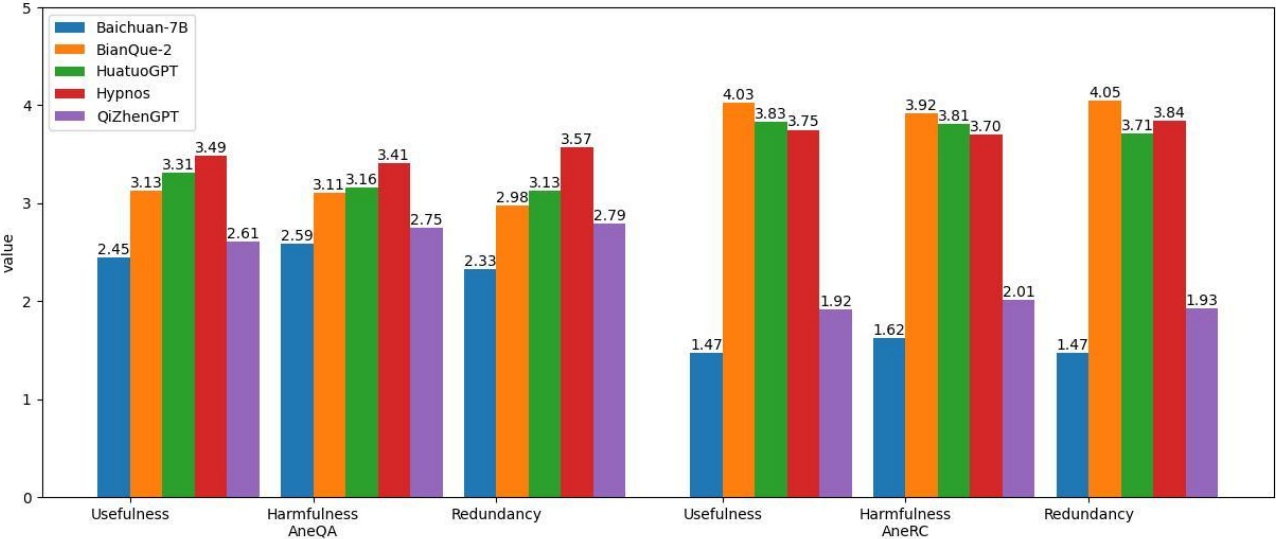}
\caption{Human Assessment Results of Hypnos and other models on the  AneRC and AneQA.}
\label{fig-3}
\end{figure*}

\subsection{Performance Comparison}
\subsubsection{Performance Using Evaluation Metrics}
Table \ref{tab:1} presents the automatic evaluation scores of the medical Language Models on the AneQA. It can be seen that our method obtained the best on most evaluation criteria. This demonstrates that our Hypnos has better ability to provide accurate answers to patients'/doctors' queries. The scores of Hypnos in Distinct-1/2 are not that high, partly because the diversity of the training data is also low. %
Table \ref{tab:2} demonstrates the performance of all methods on the single-choice question dataset AneCQ. Hypnos still obtains the best, demonstrating its ability to master Anesthesia knowledge.

\begin{figure*}[t]
\centering
\includegraphics[width=0.6\textwidth]{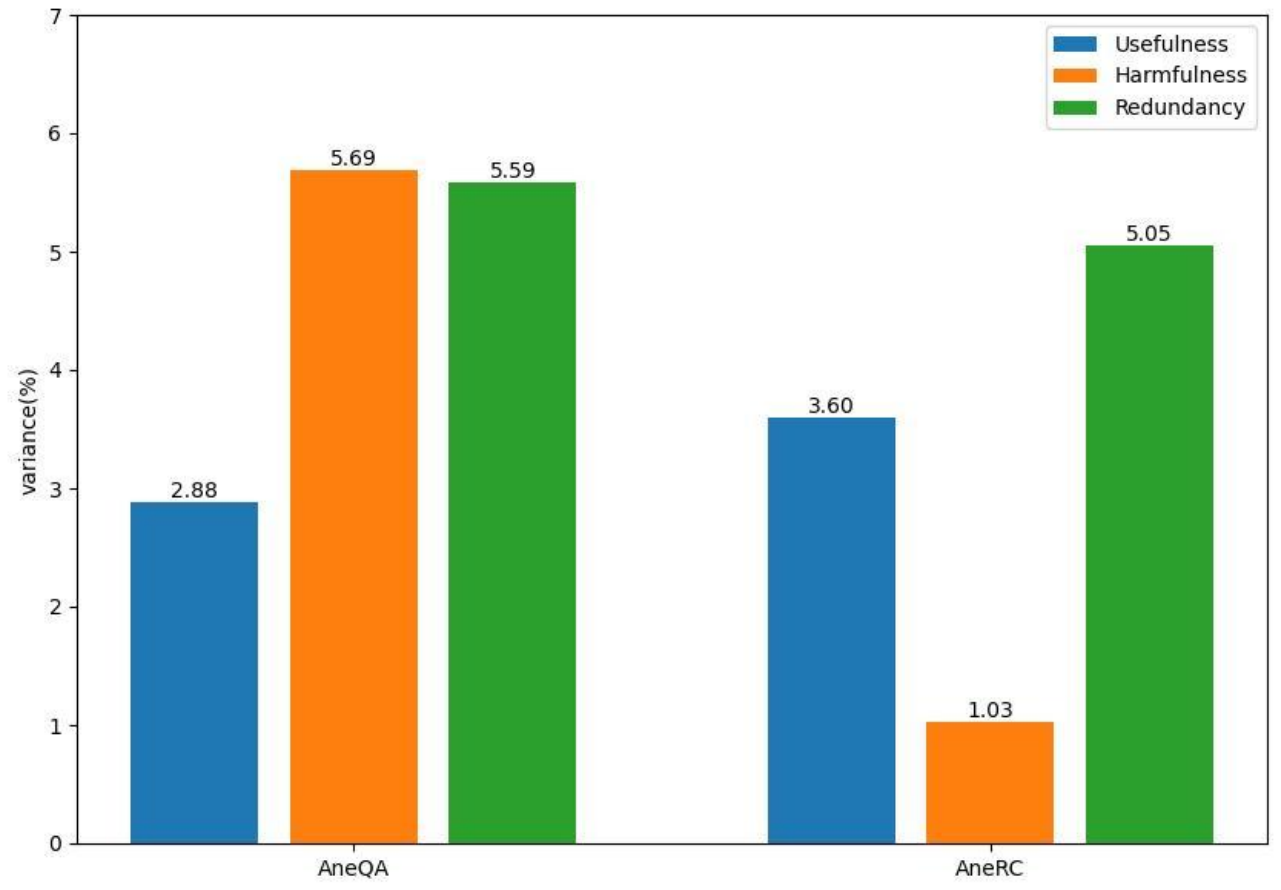}
\caption{Manual evaluation variance for each indicator on the dataset AneRC and AneQA.}
\label{fig-4}
\end{figure*}

\subsubsection{Subjective Evaluation}
Human evaluation is conducted by ranking model responses. To ensure the fairness of the experiment, all answers are presented to the doctors in a random order.%label the model responses using identifiers such as "model\_A", "model\_B", "model\_C", etc. Simultaneously, the labels assigned to the model's responses are intentionally shuffled according to specific rules. 
%听不懂
The model responses are ranked based on three aspects: usefulness, harm, and redundancy. A better model response will have a higher ranking (higher usefulness, higher ranking, lower harm, higher ranking, lower redundancy, higher ranking).

%这里是什么意思？
As shown in Figure \ref{fig-3} %As shown in \ref{tab:3}
, Hypnos achieves the highest scores in basic knowledge-based question-answering, but there is a certain gap in case analysis. The model's ability to handle long-text questions is still somewhat lacking. The reason could be that the model generates a majority of the data, lacking corresponding real cases. Further experimental studies will be conducted to investigate this issue in the future.
 
Since the evaluation is conducted using a sorting method, scores are obtained for each model in every sorting.Consequently, the variance for each model is calculated, and the variances of all models are averaged to determine the final variance for each indicator.
As shown in Figure \ref{fig-4}, the variance of each indicator on each test set is relatively small, indicating that manual evaluation has high credibility. %这里是什么意思？ because it？

\begin{table*}
  \centering
  \caption{GPT-4 Evaluation Results of Hypnos and other models on the anesthesia test set AneRC and AneQA}
  \resizebox{1\textwidth}{!}{%
  \begin{tabular}{cccccc}
    \toprule
     Dataset& metrics & Hypnos(our) vs HuatuoGPT & Hypnos(our) vs BianQue-2 & Hypnos(our) vs QiZhenGPT & Hypnos(our) vs Baichuan-7B \\
    \midrule
     &Usefulness & 54vs46 & 56vs44 & 58vs42 & 76vs24 \\
    AneQA &Harmfulness & 64vs36 & 66vs34  & 56vs44& 62vs36  \\
     &Redundancy & 68vs32 & 80vs20 &54vs46 & 54vs46  \\
     \midrule
     &Usefulness & 48vs52 & 56vs44 & 74vs26 & 84vs16 \\
     AneRC &Harmfulness & 54vs46& 60vs40& 56vs44& 62vs38  \\
     &Redundancy & 66vs34 & 46vs54 &70vs30& 84vs16\\
    \bottomrule
  \end{tabular}
  } 
  \label{tab:4}
\end{table*}

\subsubsection{Chat-based Evaluation}
As shown in Table \ref{tab:4}, the scores that are evaluated by GPT-4 are presented. The model ``Hypnos'' still performs the best in basic question-answering and also achieves relatively high scores in case analysis.
%如表4所示，经GPT4评估，模型hypnos在基础问答中依然表现最佳，在病例分析中的得分也都相对较高。

\begin{table}
  \centering
  \caption{The proportion of valid data before and after data cleaning for Claude and GPT-3.5-turbo.}
  %\resizebox{0.5\textwidth}{!}{%
  \begin{tabular}{cccc}
    \toprule
        Dataset\_model & Clean\_Model & unclean(\%) & clean(\%) \\
    \midrule
        GPT-3.5-turbo & Claude & 75.00 & \textbf{86.00}  \\
        Claude & GPT-3.5-turbo & 86.00 & \textbf{93.00}   \\
    \bottomrule
  \end{tabular}
  %}
  \label{tab:5}
\end{table}

\subsection{Ablation Experiments}
\subsubsection{Cross-filtering Influence}
Claude and GPT-3.5-turbo possess a certain level of evaluation capability; using them to assess the quality of data is a feasible approach. However, due to their limitations in capability, the evaluation scores may not be entirely accurate. To verify the effectiveness of our cross-filtering strategy, 100 pieces of data were randomly selected before and after cleaning. These two sets of data were handed over to anesthesia medical students to assess the proportion of effective data. As shown in Table \ref{tab:5}, the proportion of effective data increased after the model's cleaning process. Because many factors can influence the model's evaluation \cite{wang2023large}, it cannot guarantee that the model assesses the data from the five required aspects.

\begin{table*}
  \centering
   \caption{Based on the Llama model, training was conducted on generated data with varying degrees of cleaning, yielding automatic evaluation results. The numbers within the parentheses represent discarding data with scores equal to or below those values, while "real" denotes actual anesthesia data.}
  \resizebox{\linewidth}{!}{%
  %\resizebox{1\textwidth}{!}{%
  \begin{tabular}{cccccccccccc}
    \toprule
    Model & BLEU-1 & BLEU-2 & BLEU-3 & BLEU-4 & GLEU & ROUGE-1 & ROUGE-2 & ROUGE-L & Distinct-1 & Distinct-2 & AneCQ\\
    \midrule
               Llama(uncleared) & 17.00 & 6.12  & 2.50 & 1.26 & 8.00  & 20.32& 3.52 & 13.00 & \textbf{57.90} & \textbf{79.60} &23.1  \\
                Llama(5) & 19.27& 6.75 & 2.68 & 1.31 & 8.52 & 20.33 & 3.43 & 13.00 &   54.34  & 77.16 & 21.83 \\
    Llama(6) & 19.64 & 6.90 & 2.72 & 1.35 & 8.66 & 20.50 &3.50  & 13.25 & 54.25 & 77.28&22.97 \\
               Llama(7)& \textbf{20.49} & \textbf{7.00} & 2.79 &1.35 & 8.68 & 20.49 & 3.60  & 13.22 & 52.59 & 75.78 &22.42 \\
               Llama(6+real) &19.44 & \textbf{7.00} & \textbf{2.93} & \textbf{1.46} &\textbf{8.84} &\textbf{20.83}&\textbf{4.21} & \textbf{14.03} & 53.90  & 76.20 &25.17 \\
    \bottomrule
  \end{tabular}
  }
  \label{tab:6}
\end{table*}

\subsubsection{Data Influence} 
Following the evaluation, the anesthesia dataset was categorized into four levels: the uncleaned dataset, the dataset with scores of 5 or lower removed, the dataset with scores of 6 or lower removed, and the dataset with scores of 7 or lower removed. 
%Use these 4 kinds of data to fine-tune the llama\_7B model for 3 epochs, and select the model with the best automatic evaluation effect from the models trained in 3 epochs. 
Fine-tune the Llama\_7B model for 3 epochs using these four types of data, and select the model with the best automatic evaluation performance among the models trained for 3 epochs.
%As shown in Table \ref{tab:6}, the indicators for computing similarity, such as BLEU and Rouge, gradually increase with the increase of cleaned data, while the indicators for calculating diversity, such as Distinct, gradually decrease. 
As shown in Table \ref{tab:6}, the similarity metrics, such as BLEU and Rouge, gradually increase as the cleaning evaluation scores increase, while the diversity metrics, such as Distinct, gradually decrease. When the scores removed reached 7 points, the metrics did not show a significant increase. As indicated in Table \ref{tab:6}, the model trained on the dataset cleaned of scores 7 or below showed a decline in performance on the AneCQ test set.
Thus, our final decision was to integrate the data by excluding entries with scores of 6 or lower and incorporating real data. Fusion data demonstrated the best performance on both the AneQA and AneCQ test sets. Therefore, we opted for fusion data as our final dataset. 

\begin{table*}
  \centering
    \caption{Automated evaluation results of models trained with two different initialization methods on AneQA: initialization with UTF-8 character embeddings and random initialization.}
  \resizebox{1\textwidth}{!}{%
  \begin{tabular}{cccccccccccc}
    \toprule
     Model & BLEU-1 & BLEU-2 & BLEU-3 & BLEU-4 & GLEU & ROUGE-1 & ROUGE-2 & ROUGE-L & Distinct-1 & Distinct-2 \\
    \midrule
    Llama\_7B\_expand(utf-8) & \textbf{19.53} & \textbf{7.06}& \textbf{3.00} & \textbf{1.56} & \textbf{8.92} & \textbf{21.06} & \textbf{4.35} & \textbf{13.94} & 52.48 & \textbf{75.19}  \\
    Llama\_7B\_expand(random)& 18.64 & 6.60 & 2.80 & 1.46 & 8.50 & 20.40 & 4.06 & 13.56 &   \textbf{52.72}  & 74.00 \\    
    \bottomrule
  \end{tabular}
  }
  \label{tab:7}
\end{table*}

\subsubsection{Embedding Initialization Influence}
Using model parameters to initialize new word vectors can accelerate the convergence speed of the model within a certain amount of training. As shown in Table \ref{tab:7}, initializing with the model’s own parameters improves the scores of automated evaluation metrics compared to random initialization. This method may preserve more of the original model's syntax and semantic information.

\begin{table*}
  \centering
   \caption{The automated evaluation scores of models, both fine-tuned with general medical instruction data (Hypnos) and those not fine-tuned, on the AneQA test set."}
  \resizebox{1\textwidth}{!}{%
  \begin{tabular}{ccccccccccc}
    \toprule
     Model & BLEU-1 & BLEU-2 & BLEU-3 & BLEU-4 & GLEU & ROUGE-1 & ROUGE-2 & ROUGE-L & Distinct-1 & Distinct-2 \\
    \midrule
    Llama\_expand & \textbf{19.53} & 7.06& 3.00 & 1.56 & 8.92 & 21.06 & 4.35 & 13.94 & 52.48 & 75.19  \\
    Hypnos(our) &19.34&\textbf{7.17} & \textbf{3.1}  & \textbf{1.64}& \textbf{9.15} & \textbf{21.73} & \textbf{4.69} & \textbf{14.79}  &54.31 & 76.94  \\
     \bottomrule
  \end{tabular}
  }
  \label{tab:9}
\end{table*}

\subsubsection{General-to-specific Strategy Influence}
As shown in Table \ref{tab:9}, the anesthesia-specialized large model trained on general language corpus performs better in automated evaluation metrics. The training of general medical instruction data provides a supplementary effect on the model's capabilities in the field of anesthesia. General medical knowledge can better assist the model in comprehending the terminology and concepts within the medical context. This aids the model in understanding anesthesia domain instructions more accurately and capturing semantic correlations more effectively. Anesthesia medicine is closely related to general medicine, and the model learns certain common medical and biological knowledge, which may also be applicable in the field of anesthesia.

% 通用医疗指令数据的训练对模型在麻醉领域的能力有一定的补充作用，通用医疗知识可以更好的帮助模型理解医学上下文中的术语和概念。这有助于模型更准确地理解麻醉领域的指令，并更好地捕捉语义关联。麻醉医学与通用医疗密切相关，模型学习到一些通用的医学和生物学知识，这些知识可能在麻醉领域同样适用。

\section{Conclusion}
This paper presents the first Chinese Anesthesia Large Language Model: Hypnos. Two types of useful strategies: 1) cross-filtering strategy to obtain high-quality data from existing LLMs and 2) general-to-specific training strategy to fully utilize general and specific medical data are introduced to obtain an LLM of the specific medical field. The experimental results on the anesthesia dataset demonstrate our Hypnos' competitiveness and our proposed strategies' usefulness. Learning an LLM in Anesthesiology (or a specific medical field) is non-trivial, and we hope the studies of Hypnos will benefit the development of LLMs for a specific medical field.

\section{Limitation}
The pre-trained model used in this paper is trained based on a large corpus, so the correctness of the corpus data cannot be fully guaranteed. There may exist biases, errors, and incompleteness in the training process. This model is for reference only and cannot guarantee the accuracy and reliability of its answers. We do not bear any responsibility for the results generated by using the pre-trained model or any loss caused by using the pre-trained model. Users should verify the correctness of the model's answers on their own when using the model.
\begin{comment}
\section{Acknowledgments}
We would like to express our deepest gratitude to teachers Hua Jin and Jun Peng, as well as their students, for their contributions to our anesthesia manual assessment. Special thanks go to [Name] at [Institution] for providing [resource or assistance].We acknowledge the support of [Funding Agency] for funding this research project.We are grateful to our colleagues at [Institution] for their valuable feedback and discussions.The authors would like to thank the anonymous reviewers for their insightful comments.
\end{comment}

\bibliographystyle{IEEEtran}
%\bibliography{New_IEEEtran_how-to}
\begin{CJK}{UTF8}{gbsn}
{
% Generated by IEEEtran.bst, version: 1.14 (2015/08/26)

}
\end{CJK}

\end{document}